\documentclass[runningheads,a4paper]{llncs}

\usepackage{amssymb}
\usepackage{graphicx}
\usepackage[justification=centering]{caption}
\usepackage{subcaption}
\usepackage{xcolor}
\usepackage{float}
\usepackage{url}
\usepackage{amsmath}
\usepackage{booktabs} 
\usepackage{tabularx}
\usepackage[english]{babel}
\usepackage[normalem]{ulem}
\usepackage{cite}
\useunder{\uline}{\ul}{}
\newcommand{\keywords}[1]{\par\addvspace\baselineskip
\noindent\keywordname\enspace\ignorespaces#1}

\begin{document}

\mainmatter  

\title{A Trainable Reconciliation Method \\ for Hierarchical Time-Series}
\titlerunning{A Trainable Reconciliation Method for Hierarchical Time-Series}

%
%
\author{Davide Burba \and Trista Chen}
\authorrunning{Davide Burba \and Trista Chen}

\institute{Inventec Corporation - AI Center, Taiwan}

%
%

\toctitle{A Trainable Reconciliation Method for Hierarchical Time-Series}
\tocauthor{Authors' Instructions}
\maketitle

\begin{abstract}
In numerous applications, it is required to produce forecasts for multiple time-series at different hierarchy levels. An obvious example is given by the supply chain in which demand forecasting may be needed at a store, city, or country level. The independent forecasts typically do not add up properly because of the hierarchical constraints, so a reconciliation step is needed. In this paper, we propose a new general, flexible, and easy-to-implement reconciliation strategy based on an encoder-decoder neural network. By testing our method on four real-world datasets, we show that it can consistently reach or surpass the performance of existing methods in the reconciliation setting.

\end{abstract}

\section{Introduction}

\begin{figure}[h]
\centering
\begin{subfigure}{\linewidth}
    \centering
    \includegraphics[height=4.0cm]{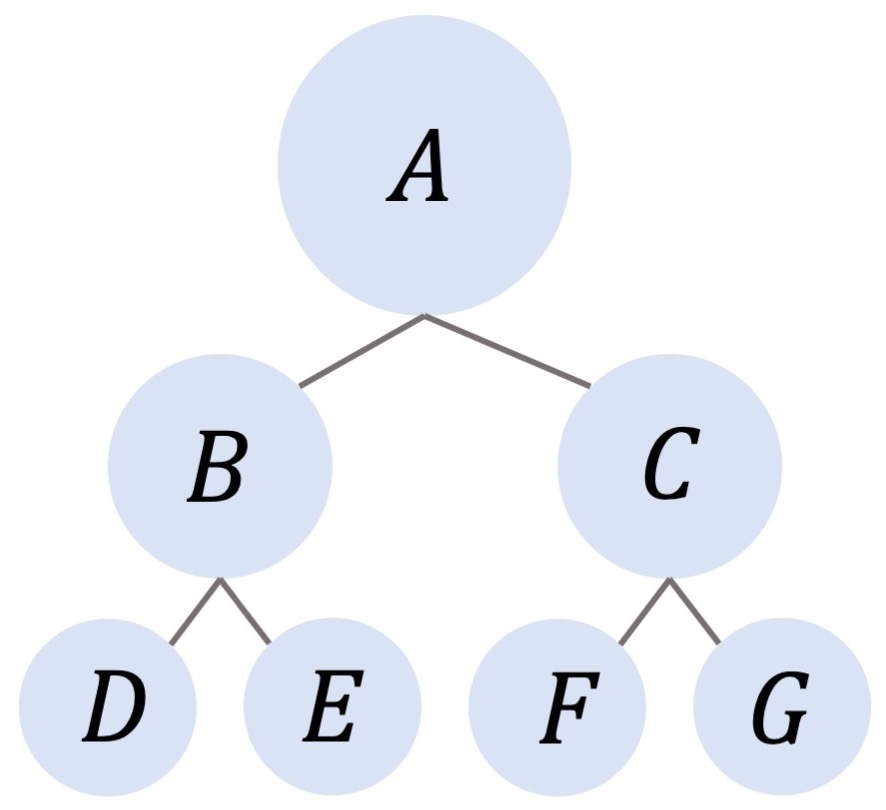}
    \caption{Time-series hierarchy}
\end{subfigure}

\begin{subfigure}{\linewidth}
    \centering
    \includegraphics[width=\linewidth]{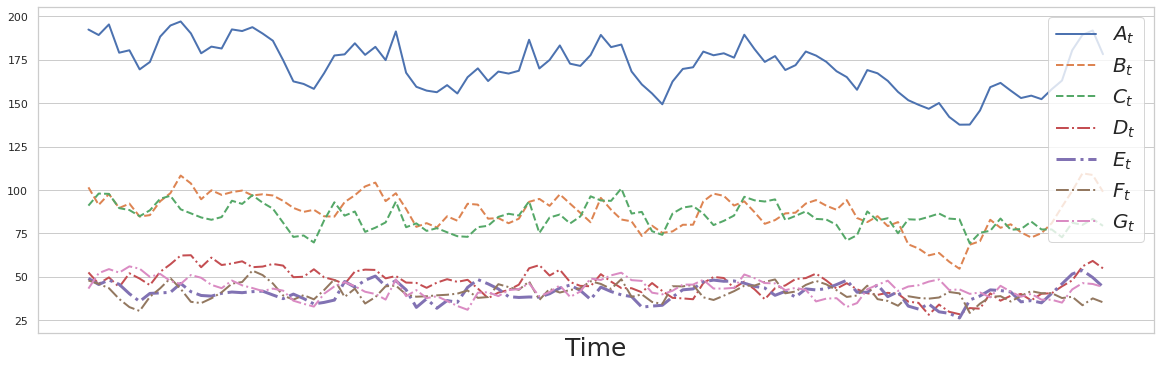}
    \caption{Time-series values}
\end{subfigure}
\caption{Example of a hierarchical time-series: \\ 
$A_t = B_t + C_t $ \\
$B_t = D_t + E_t $ \\
$C_t = F_t + G_t $ }
\label{fig:ex:a}
\end{figure}

A hierarchical time-series is a collection of time-varying observations organized in a hierarchical structure. The problem of forecasting hierarchical time-series often appears in business and economics, where time-varying quantities need to be predicted at different granularity levels. 
For instance, in the supply chain, forecasts of the demand may be required at a country, city, or store level to organize the logistic. A synthetic example of a hierarchical time-series is shown in Figure \ref{fig:ex:a}.

Predictions for hierarchical time-series are typically built in two independent stages. First, forecasts are produced for all or some of the time-series. Then, the forecasts are reconciled to enforce the hierarchy constraints.
The main methods for reconciling time-series predictions are known as ``bottom-up", ``top-down", ``optimal reconciliation", and ``trace minimization". These methods can be expressed in a common framework \cite{hyndman2011optimal}, and at the moment there is no consensus about their performance. 
A common drawback of existing methods is that they are not flexible, meaning that they do not allow for a specific metric to be optimized. However, in real-world forecasting models, it is typically required to minimize a given metric, e.g. the Mean Absolute Scaled Error (MASE) or the Mean Logarithm of Absolute Error (MLAE), and the modeling choices depend on the chosen metric.

With the rise of deep learning in the past years, some attempts have been proposed to improve performances and overcome current limitations in the reconciliation setting \cite{mishchenko2019self,mancuso2020machine}. These methods exploit the hierarchical structure by imposing soft constraints in the loss function to regularize the training process, improve forecasting performances, and tighten the reconciliation gaps. However, they cannot guarantee an exact reconciliation, i.e. the hierarchy constraints are not satisfied.

In this work, we propose a new exact methodology to reconcile hierarchical time-series forecasts based on an encoder-decoder neural network. The encoder is a trainable neural network that takes as input the independent forecasts and outputs the bottom-level reconciled forecasts. The decoder is a fixed matrix which reconstructs exactly the forecasts at all levels using the bottom level encoded predictions.
Our method includes and generalizes the representation space of existing methods, is extremely flexible, and is easy to implement.
We apply it to four real-world datasets, and we show that it consistently achieves a better or equal performance than the existing reconciliation methods.

To summarize, the main contributions of this paper are: (1) we propose a new easy-to-implement reconciliation method which is more general and flexible than existing approaches, (2) we empirically show that this method improves over the state-of-the-art by testing it on multiple heterogeneous datasets.

The rest of the paper is organized as follows. In Section \ref{related_works} we briefly describe existing reconciliation methods. In Section \ref{proposed_method} we describe the proposed method and in Section \ref{implementation} we show our implementation. In Section \ref{datasets} we describe the datasets used in the experiments and in Section \ref{results} we show the results of the experiments. We conclude in Section \ref{conclusion}.

\section{Related Works}
\label{related_works}

In this Section we briefly introduce the main existing reconciliation methods, i.e. bottom-up, top-down, optimal reconciliation, and trace minimization.

\vspace{-3mm}
\subsubsection{Bottom-Up} 
The Bottom-Up (BU) \cite{hyndman2011optimal} approach consists in producing forecasts only for the bottom level of the hierarchy and then aggregating them to the upper levels. It is the simplest reconciliation method.

\vspace{-3mm}
\subsubsection{Top-Down} 
The Top-Down approach consists of making predictions at the top level of the hierarchy and distributing them to the lower levels. Different splitting strategies correspond to different Top-Down methods. The most popular ones are Top-Down Historical Proportions (TDHP) \cite{gross1990disaggregation} which computes the splits based on the historical time-series proportions, and Top-Down Forecasted Proportions (TDFP) \cite{athanasopoulos2009hierarchical} which, at each forecasted time-step, iteratively computes the forecasted proportions while descending the hierarchy tree. Note that TDFP requires predictions at all levels of the hierarchy to compute the proportions.

TD and BU approaches may be mixed in the Middle-Out (MO) approach. With MO, we first select an intermediate level of the hierarchy; then, predictions are computed via BU for upper levels and TD for lower levels.

\vspace{-3mm}
\subsubsection{Optimal-Combination} 
The Optimal Combination (OC) method \cite{hyndman2011optimal} requires to produce forecasts for all the time-series at all levels of the hierarchy. The predictions are then mixed via a constrained linear combination to produce the reconciled predictions. The values of the linear coefficients are based on an estimate of the covariance matrix of the incoherence errors.

\vspace{-3mm}
\subsubsection{Trace-Minimization} 
In \cite{wickramasuriya2019optimal} it is shown how the covariance matrix of the incoherence errors for the OC method is non-identifiable, and therefore impossible to estimate. 
The authors propose instead the Trace-Minimization (TM) method \cite{wickramasuriya2019optimal}, a constrained linear combination method whose estimate is based on the covariance matrix of the forecast errors. TM estimate has the property of minimizing the mean squared error of the coherent forecasts under the constraint of unbiasedness.

\vspace{5mm}
All the methods described above can be framed in a common framework \cite{hyndman2011optimal}. Let us denote by \small{$N$} the total number of time-series at all levels and by \small{$M$} the number of time-series at the bottom level. Let $\mathbf{\hat{y}}_t$ be the vector containing the individual time-series predictions at time $t$ and $\mathbf{\tilde{y}}_t$ the vector with the  corresponding reconciled predictions. Then, the methods described above can be expressed as:

\begin{equation}
    \mathbf{\tilde{y}}_t = SP\mathbf{\hat{y}}_t
\end{equation}

where $P$ is a method-specific \small{$(M\times N)$} matrix which maps the individual predictions to the bottom-level reconciled predictions, and $S$ is the hierarchy-specific \small{$(N\times M)$} matrix which reconstructs the predictions at all levels of the hierarchy by summing the corresponding bottom-level values.

As an example, let us consider the hierarchy illustrated in Figure \ref{fig:ex:a}. If we set $\mathbf{\hat{y}}_t=[\hat{A}_t,..,\hat{G}t]^T$ and $\mathbf{\tilde{y}}_t=[\tilde{A}_t,..,\tilde{G}_t]^T$, then the BU approach corresponds to the following matrices:

\begin{equation*}
P^{(BU)} = 
\begin{bmatrix}
{\color{gray}0} & {\color{gray}0} & {\color{gray}0} & 1 & {\color{gray}0} & {\color{gray}0} & {\color{gray}0} \\
{\color{gray}0} & {\color{gray}0} & {\color{gray}0} & {\color{gray}0} & 1 & {\color{gray}0} & {\color{gray}0} \\
{\color{gray}0} & {\color{gray}0} & {\color{gray}0} & {\color{gray}0} & {\color{gray}0}  & 1 & {\color{gray}0} \\
{\color{gray}0} & {\color{gray}0} & {\color{gray}0} & {\color{gray}0} & {\color{gray}0} & {\color{gray}0} & 1 \\
\end{bmatrix}
\hspace{5mm}
S = 
\begin{bmatrix}
1 & 1 & 1 & 1 \\
1 & 1 & {\color{gray}0} & {\color{gray}0} \\
{\color{gray}0} & {\color{gray}0} & 1 & 1 \\
1 & {\color{gray}0} & {\color{gray}0} & {\color{gray}0} \\
{\color{gray}0} & 1 & {\color{gray}0} & {\color{gray}0} \\
{\color{gray}0} & {\color{gray}0} & 1 & {\color{gray}0} \\
{\color{gray}0} & {\color{gray}0} & {\color{gray}0} & 1 \\
\end{bmatrix}
\end{equation*}

and therefore:

\begin{equation*}
\mathbf{\tilde{y}}_t = SP^{(BU)}
\begin{bmatrix}
\hat{A}_t  \\
\hat{B}_t  \\
\hat{C}_t  \\
\hat{D}_t  \\
\hat{E}_t  \\
\hat{F}_t  \\
\hat{G}_t  \\
\end{bmatrix}
= 
\begin{bmatrix}
\hat{D}_t + \hat{E}_t + \hat{F}_t + \hat{G}_t  \\
\hat{D}_t + \hat{E}_t  \\
\hat{F}_t + \hat{G}_t  \\
\hat{D}_t  \\
\hat{E}_t  \\
\hat{F}_t  \\
\hat{G}_t  \\
\end{bmatrix}
\end{equation*}

For the TD approach, the $P$ matrix is filled with zeros except for the first column, which sums-up to one. For the optimal combination and trace-minimization methods, $P$ is a dense matrix.

\section{Proposed Method}
\label{proposed_method}

\begin{figure}[h]
\begin{center}
\includegraphics[width=\textwidth]{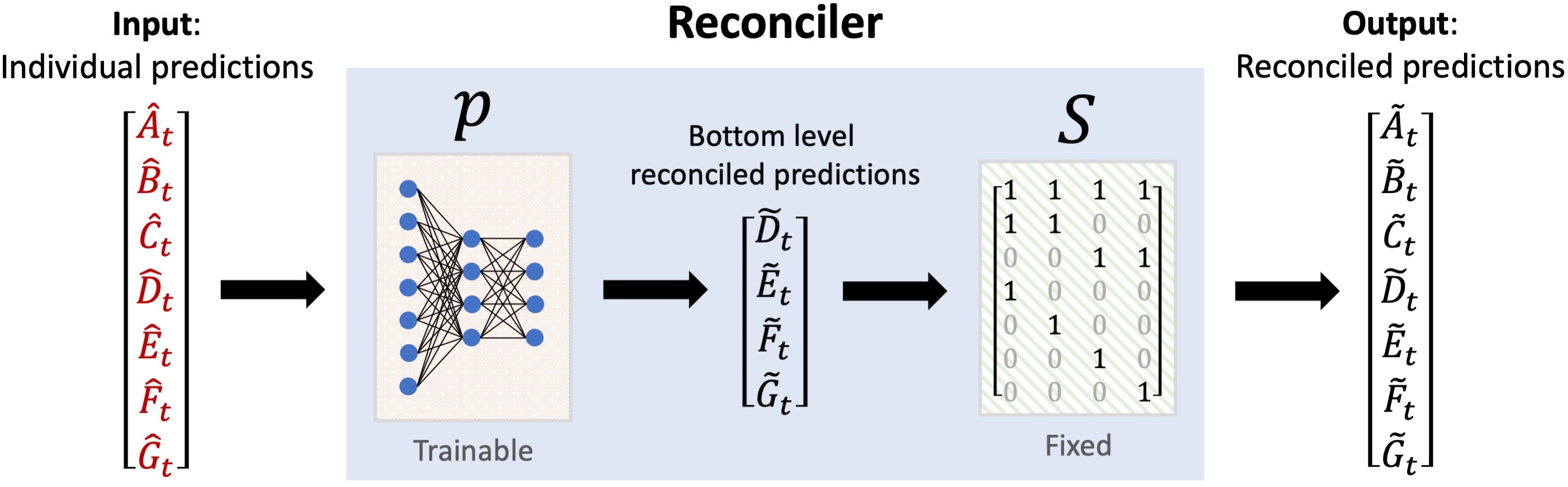}
\caption{Trainable reconciler structure (refer to hierarchy in Figure \ref{fig:ex:a})}
\label{fig:example_hts}
\end{center}
\end{figure}

In our approach we propose to generalize the \small{$(M\times N)$} matrix $P$ described in Section \ref{related_works} with a generic function $p:\mathbb{R}^N\rightarrow\mathbb{R}^M$ represented via a neural network. 
We can think of the proposed method as a simple encoder-decoder neural network, where the network $p$ represents the (trainable) encoder and the matrix $S$ represents the (fixed) decoder.
The encoder maps the input predictions to the bottom level reconciled predictions, and the decoder takes the latter as input and reconstructs the predictions at all levels.
A representation of our method is given in Figure \ref{fig:example_hts}.

This approach has two main theoretical advantages: generalization and flexibility. The generalization advantage refers to the wide representation space of our model, which includes the one described in Section \ref{related_works} and extends it, allowing for non-linearities.
The flexibility advantage refers to the fact that our approach allows us to target specific performance goals via the choice of a suitable loss function in the training phase. Existing methods are simple heuristics (bottom-up, top-down) or minimize the estimated coefficients' errors under different assumptions (optimal-combinations, trace-minimization). Our method instead allows us to target different metrics. For instance, if our target is the Mean Logarithm of Absolute Error (MLAE), we can use the MLAE as a loss function. Or, if we are especially interested in a specific level of the hierarchy, we can assign a higher weight to the corresponding loss terms. Similarly, if errors' importance depends on a scale or is asymmetric, we can change the loss function accordingly so to optimize for our target. 

Last but not least, our model has the practical advantage to be easy-to-implement and accessible to the wide and fast-growing deep learning community, as opposed to complex statistical models such as optimal combination or trace minimization. Furthermore, if the forecasting models can be expressed in a deep-learning framework, our approach allows us to stack the reconciliation network on top of them and simultaneously train the reconciler and fine-tune the forecasting models.

\section{Implementation}
\label{implementation}

In this Section we describe how we implement our trainable reconciliation method to perform the experiments. We use \textsc{Sklearn} \cite{scikit-learn} and \textsc{LightGBM} \cite{ke2017lightgbm} libraries for the forecasting models, and \textsc{PyTorch} \cite{paszke2017automatic} framework for our reconciliation method. We develop our analysis in \textsc{Python}, and we use \textsc{R} to fit the Optimal-Combination and Trace-Minimization methods exploiting the \textsc{HTS} \cite{hyndman2015hts} package.

\subsection{Time-Series Prediction}
\label{sec:ts_pred}
For simplicity, we only consider the task of forecasting the one-step-ahead time-series values.
To train the forecasting models we consider two alternative strategies, i.e. estimating an individual model for each time-series, or a global model for all of them. 
For the individual strategy, we consider linear autoregressive models ($AR(p)$), taking the lagged time-series values as input. For the datasets which present a high number of time-series ($>500$), we estimate a single global predictive model to forecast all the time-series. This approach allows us to exploit the time-series similarities and estimating a more complex model \cite{salinas2020deepar}. We consider the Light Gradient Boosting (LightGBM) model, taking as input the scaled lagged values, time-series specific features (such as the level in the hierarchy), and temporal features. 

For both modeling strategies, we choose the best hyperparameters combination by performing ten-folds blocked cross-validation, i.e. with validation sets belonging to the same time-window. This validation technique was shown to be effective for time-series tasks \cite{bergmeir2012use,bergmeir2018note}. 
For the individual strategy with $AR(p)$, we perform a grid search to optimize the number of predictive lagged values $p$.
For the global strategy with LighGBM, we perform a grid search to optimize the \textit{number of leaves} and \textit{minimum number of observations in each leaf} hyperparameters, and we keep the default values for all the others. 
Once the best configuration is found, we keep the ten models trained during cross-validation in order to validate the reconciler parameters, as shown in Section \ref{valid_strategy}.

\subsection{Reconciliation}

\begin{figure}[h]
\centering
\begin{subfigure}{.4\linewidth}
    \centering
    \includegraphics[height=3.0cm]{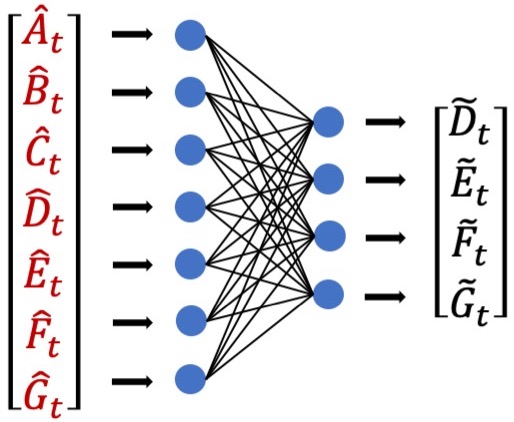}
    \label{fig:shrink:a}
    \caption{Fully connected}
\end{subfigure}%
\begin{subfigure}{.4\linewidth}
    \centering
    \includegraphics[height=3.0cm]{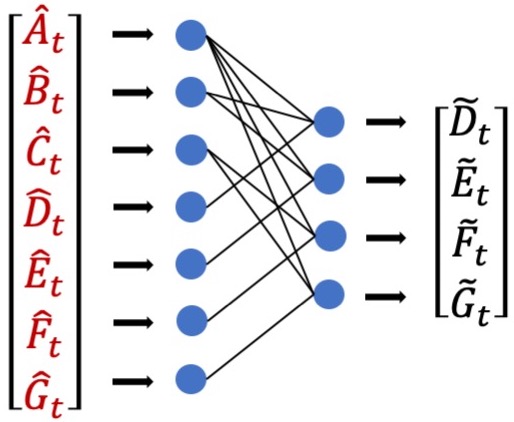}
    \label{fig:shrink:b}
    \caption{Shrunk}
\end{subfigure}%
\caption{Encoder architecture alternatives, without hidden layers for ease of visualization. Refer to hierarchy in Figure \ref{fig:ex:a}.}
\label{fig:shrink}
\end{figure}

For the reconciler's encoder, we consider feed-forward networks with ReLU activation functions and we set the size of the output layers to the number of bottom level time-series. 
We consider two architecture alternatives. The first is a standard fully connected network. The second is a ``shrunk" version of a fully connected network such that the output for a given bottom-level time-series depends only on forecasts for itself and his parents in the hierarchy at all levels. As an example, consider the hierarchy in Figure \ref{fig:ex:a}. Let us consider the reconciled predictions for bottom level time-series $\tilde{D}_t$. In the fully connected case, we have that $\tilde{D}_t$ is influenced by forecasts for all the time series at all levels: $\tilde{D}_t = f(\hat{A}_t,...,\hat{G}_t)$. In the shrunk case instead, $\tilde{D}_t$ depends only on the forecast for itself, its parent and its grand-parent: $\tilde{D}_t = f(\hat{A}_t,\hat{B}_t,\hat{D}_t)$.
The ``shrunk" version does not have the same representative power of the fully connected one. However note that it still includes bottom-up, top-down, and middle-out representation spaces, and any mix of them. Furthermore, for a given number of hierarchy levels, the number of trainable parameters grows with the number of time-series quadratically in the fully connected case and linearly in the shrunk case. To reduce the overfitting risk, we consider the shrunk architecture when the number of time-series is greater than ten times their length. However, the shrunk option can be generally considered as a tunable hyperparameter.
 An illustration of the two encoder architectures is given in Figure \ref{fig:shrink}.

Applying the model directly to the raw time-series presents a challenge: hierarchical time-series naturally exhibit large amplitude differences, which could compromise the model training since neural networks are sensitive to the scale of the data \cite{589532}. To deal with the scale problem, we perform a normalization step. We divide each time-series by a scaling factor before feeding the data to the encoder, and we perform the inverse scaling on the encoded output. The scaling factor is set to one plus the time-series average computed in the training period.

Given the short size of the training samples, learning a multi-dimensional transformation could be challenging for the network. To simplify the training process, we want to initialize the network to the BU method, meaning that the untrained encoder should map the raw predictions to the raw bottom-level predictions.  We explore two different ways to achieve this, i.e. using residual connections \cite{he2016deep} or using a custom initialization of the network weights. We choose the second option due to its slightly higher performance, and we refer to this method as ``bottom-up initialization". As an example, consider the two architectures shown in Figure \ref{fig:shrink}. With both these architectures we initialize the weights connecting each bottom level input to the corresponding output to one (i.e. the connections $\hat{D}_t \rightarrow {\tilde{D}_t},\hat{E}_t \rightarrow {\tilde{E}_t},\hat{F}_t \rightarrow {\tilde{F}_t},\hat{G}_t \rightarrow {\tilde{G}_t}$) and all the other weights and biases to zero. 
This technique can be trivially extended to multiple layers, assuming that the size of the hidden layers is at least equal to the output size.

To improve the stability of the model, we ensemble ten networks and we take the average prediction.

\subsection{Metrics}

To demonstrate the flexibility of our approach, we consider two evaluation metrics. The first one is the Mean Absolute Scaled Error (MASE) \cite{hyndman2006another}, which scales the predictions errors by the average in-sample na\"ive error and has the nice properties of being symmetric, stable for values close to zero, and scale-free. 
These, in general, are desirable statistical properties. However, the scale-free property sometimes may not be aligned with the business point of view. If, on the one hand, scale-dependent metrics such as Mean Absolute Error (MAE) may be misleading since the prediction error for small time-series would become irrelevant in presence of much larger ones, on the other hand, higher forecast errors typically correspond to higher costs, regardless of the time-series scale. Furthermore, the MASE is undefined or unstable for constant or almost constant time-series, respectively.
For these reasons, we also consider the Mean Logarithm of Absolute Error (MLAE). The MLAE depends on the time-series scale, but at the same time its dependency is ``reduced" via the natural logarithmic transformation $x\mapsto log(x+1)$, and is stable for constant and almost constant time-series.

We trained the models with different loss functions based on the two target metrics: the MASE loss for the MASE metric, and the MLAE loss for the MLAE metric. Note that using the performance metric as a loss function is a convenient way to optimize for our target and is proven effective in our experiments. However, there are cases in which this may not be the optimal solution. For instance, using the MASE as a loss may be unstable in case of almost constant time series; if this is the case we suggest to use a different loss or a regularized version of it, e.g. by scaling the errors by one plus the average in-sample na\"ive error.

\subsection{Validation Strategy}
\label{valid_strategy}

To choose the hyperparameters for the reconciliation model we exploit, again, ten folds blocked cross-validation. 
Specifically, we divide the training time interval into ten time-windows, and at each iteration, we perform the following steps:
\begin{enumerate}
    \item Set one time window as \textit{validation period}, and the remaining ones as \textit{training period}.
    \item Select the forecasting models trained on the \textit{training period} and use them to generate forecasts for both periods. These are the reconciler inputs in the next step.
    \item Train the reconciler with forecasts in the \textit{training period} and evaluate its performance with forecasts in the \textit{validation period}.
\end{enumerate}

Note that, for each fold, we produce the reconciler input with the forecasting models trained on the reconciler training period. One may wonder why we do not simply use the predictions of the forecasting models trained on the whole training period. This is because in doing so, the validation labels for the reconciler would be used twice, first to train the forecasting models and then to evaluate the reconciler, and this would lead to an over-optimistic evaluation.

We select the best combination of hyperparameters via random search; details about the training procedure and the hyperparameters search are shown in the Appendix.

\section{Datasets}
\label{datasets}

For our experiments we consider four real-world datasets, with an increasing complexity in the hierarchy. We denote them  as \textsc{Dairy} \cite{dairy}, \textsc{Walmart} \cite{walmart}, \textsc{Australian} \cite{australian},  and \textsc{Electricity} \cite{energy}.
A short description of the datasets is given here below, and detailed instructions on the data collection process are available at \url{https://github.com/davide-inventec/trainable-reconciler-data}. Table \ref{tab:data_stats} provides an overview of the datasets statistics.  All the datasets are scaled by the overall absolute average on the training set. 

\vspace{-3mm}
\subsubsection{Dairy}
Weekly sales of dairy products in the United States (volume in USD). The hierarchy levels are total, dairy products (milk, butter, etc.).

\vspace{-3mm}
\subsubsection{Walmart} 
Daily sales for a popular Walmart product. The hierarchy levels are total, country, store. 

\vspace{-3mm}
\subsubsection{Australian}
Quarterly time-series of domestic tourism in Australia. The hierarchy levels are total, purpose of travel (holiday, business, etc.), region (Sidney, Brisbane, etc.), destination (city, non-city).

\vspace{-3mm}
\subsubsection{Electricity}
Yearly electricity consumption in MWH in the State of New York. The hierarchy levels are total, county (Bronx, Richmond, etc.), type of consumer (Business, Residential, etc.), and provider (National grid, Central Hudson, etc.).

\begin{table}[H]
\caption{Datasets statistics.}
\label{tab:data_stats}
\vskip 0.15in
\begin{center}
\begin{small}
\begin{sc}

\begin{tabular}{lccp{0.2\textwidth}<{\centering}cc}
\toprule
Dataset   & \# levels & \# time-series & \begin{tabular}[c]{@{}c@{}}\# time-series\\  per level\end{tabular} & \begin{tabular}[c]{@{}c@{}}train\\  time-steps\end{tabular} & \begin{tabular}[c]{@{}c@{}}test\\  time-steps\end{tabular} \\ 
\midrule
Dairy       & 2         & 6              & (1,5)                                                               & 380          & 53          \\
Walmart     & 3         & 13             & (1,3,9)                                                             & 1575         & 366         \\
Australian  & 4         & 89             & (1,4,28,56)                                                         & 28           & 8           \\
Electricity & 4         & 772            & (1,60,295,416)                                                      & 36           & 12          \\ 
\bottomrule
\end{tabular}
\end{sc}
\end{small}
\end{center}
\vskip -0.1in
\end{table}

For the \textsc{Electricity} dataset we use a global forecasting model, using 13 numerical features (12 lagged values scaled by the time-series average, and the time-series average) and 6 categorical features (the month, 3 level-specific variables, the hierarchy level,  and the time-series ID) as predictors. 
For all the other datasets, we use individual forecasting models.
We use the shrunk architecture for \textsc{Electricity} and the fully connected one for all the others.

\section{Results}
\label{results}

We evaluate the performance of different methods by computing the MASE and MLAE on the test periods for each dataset. To compute the scores, we average the prediction errors for all the time-series at all hierarchy levels.
Then, we assess the significance of performance differences between our method and the \textit{best} among the others by performing paired $t$-tests at 1\% and 5\% on the corresponding scaled/logarithmic absolute errors.
In Table \ref{tab:performances} we show the performance results, highlighting \textit{significantly} better scores in bold and using one or two stars to denote p-values smaller than 5\% and 1\%, respectively. The detailed scores for each hierarchy level are shown in the Appendix.
\vspace{-2mm}

\begin{table}[h]
\caption{Performances on the test sets.}
\label{tab:performances}

\begin{subtable}{\linewidth}
\caption{MASE}
\label{tab:performances:mase}
\begin{center}
\begin{small}
\begin{sc}
\begin{tabular}{lp{0.1\textwidth}<{\centering}p{0.1\textwidth}<{\centering}p{0.1\textwidth}<{\centering}p{0.1\textwidth}<{\centering}p{0.1\textwidth}<{\centering}c}
\toprule
               & BU             & TDHP           & TDFP          & OC      & TM             & Trainable (ours) \\
\midrule
Dairy          & 1.1011         &	1.5543	     &	1.0960	     &	1.1150    &	1.1014         	&	1.1037               \hspace{4.3mm}  \\
Walmart        & 0.6608         &	0.7823	     &	0.6600	     &	0.6606	  &	0.6580	        &	0.6567      \hspace{4.3mm}  \\
Australian     & 0.6472         &	0.7181	     &	0.6918	     &	0.7602	  &	0.6301	        &	\textbf{0.5910 **}                   \\
Electricity    & 0.7041         &	0.9629	     &	0.7216	     &	3.0777	  &	0.7042	        &	\textbf{0.7038 *}    \hspace{1.1mm}  \\
\bottomrule
\end{tabular}
\end{sc}
\end{small}
\end{center}
\end{subtable}
\begin{subtable}{\linewidth}
\caption{MLAE}
\label{tab:performances:mlae}
\begin{center}
\begin{small}
\begin{sc}
\begin{tabular}{lp{0.1\textwidth}<{\centering}p{0.1\textwidth}<{\centering}p{0.1\textwidth}<{\centering}p{0.1\textwidth}<{\centering}p{0.1\textwidth}<{\centering}c}
\toprule
               & BU             & TDHP           & TDFP          & OC    & TM             & Trainable (ours) \\
\midrule
Dairy          &  0.1110           &	0.1255	&  0.1113   & 	0.1114	&   0.1112  &  0.1129            \hspace{4.4mm} \\
Walmart        &  0.1070           &	0.1169	&  0.1065   & 	0.1064	&   0.1064  &  0.1060   \hspace{4.4mm} \\
Australian     &  0.0929           &	0.1144	&  0.1045   & 	0.1018	&   0.0923  &  \textbf{0.0883 *} \hspace{1.1mm} \\
Electricity    &  0.0569           &	0.0742	&  0.0580   & 	0.0600	&   0.0575  &  \textbf{0.0567 **}               \\
\bottomrule
\end{tabular}
\end{sc}
\end{small}
\end{center}
\end{subtable}
\end{table}

For the datasets with a complex hierarchy, our method performs significantly better than all the others, while for the simpler ones the difference across different methods is generally small.  In particular, for the \textsc{Australian} and the \textsc{Electricity} datasets, our method achieves a significantly better performance than all the other methods in terms of both MASE and MLAE. For the \textsc{Walmart} dataset, our method has the highest performance, however, the difference from the second best is not statistically significant. For the \textsc{Dairy} dataset, all the methods have statistically similar performances (except for TDHP, which has larger errors). We think that this is because the \textsc{Dairy} dataset exhibits a trivial two levels hierarchy, and therefore there is not much space for improvement.

To summarize, we find that our method consistently achieves a better or equivalent performance than the best of all the other methods, and the greatest improvement corresponds to the datasets with the deepest hierarchies. Given that scores for other methods are highly variable (especially for TDHP, TDFP, and OC), we believe that this is a strong result.

\section{Conclusion}
\label{conclusion}

We have shown that a deep learning based approach can improve the forecasting accuracy over state-of-the-art methods in the time-series reconciliation setting. Our method, which is based on an encoder-decoder feed-forward neural network, generalizes the representation space of existing methods, allows us to target specific metrics via the choice of an appropriate loss function, and is easy-to-implement. By testing our model on four real-world datasets, we saw that it was able to achieve, statistically, better or equivalent performances than the best of all the other methods used in the experiments.

For what concerns future developments, we believe that our method may open multiple research directions. In particular, further exploration could be done to evaluate the possibility of jointly training the forecasting models and the reconciliation network.
Besides, since our custom initialization converges to a solution that is probably close to the bottom-up method, it would be interesting to investigate the effect of initialization strategies on different architectures.

\section*{Appendix}

\subsection*{Reconciler Hyperparameters Search}

We train all the models via backpropagation, using the AdamW \cite{loshchilov2017decoupled} optimizer and a batch size equal to 128 . For the fully connected networks, we set the size of the hidden layers to the number of bottom level time-series. For shrunk networks, we set the size of the hidden layers for each sub-network relative to each bottom level time-series to 8, and therefore the number of nodes in each hidden layer is equal to 8 times the number of bottom level time-series.

We optimize the reconciler hyperparameters by evaluating 100 random combinations via the procedure described in Section \ref{valid_strategy}. Combinations were uniformly sampled from the following values:
\vspace{-1mm}
\begin{itemize}
\item \textit{dropout} : $[0,0.1,0.2]$
\item \textit{learning rate} : $[1^{-3},1^{-4},1^{-5}]$
\item \textit{weight decay} : $[1^{-1},3^{-2},1^{-2}]$
\item \textit{number of training epochs} : $[50,100,200,500]$
\item \textit{number of hidden layers} : $[0,1,2,3]$
\end{itemize}

\subsection*{Results by Hierarchy Level}

In Table \ref{tab:performances_by_level} we show the performances on the test set for each dataset, at each level of the hierarchy.  Level 0 denotes the top level.

\begin{table}[H]
\caption{Performances on the test sets by hierarchy level.}
\label{tab:performances_by_level}

\vspace{1mm}
\begin{subtable}{\linewidth}
\begin{center}
\begin{small}
\begin{sc}
\begin{tabular}{lp{0.1\textwidth}<{\centering}p{0.1\textwidth}<{\centering}p{0.1\textwidth}<{\centering}p{0.1\textwidth}<{\centering}p{0.1\textwidth}<{\centering}c}
\toprule
\textbf{DAIRY}               & BU             & TDHP           & TDFP          & OC    & TM             & Trainable (ours) \\
\midrule
mase - level 0 \hspace{3mm}    &   1.0865   &   1.6276   &   1.0776   &   1.1014   &   1.0868   &   1.0840  \\
mase - level 1   &   1.1740   &   1.1878   &   1.1878   &   1.1829   &   1.1748   &   1.2023  \\
\midrule
mlae - level 0   &   0.0837   &   0.1007   &   0.0836   &   0.0839   &   0.0840   &   0.0844  \\
mlae - level 1   &   0.2475   &   0.2497   &   0.2497   &   0.2488   &   0.2475   &   0.2553  \\
\bottomrule
\end{tabular}
\end{sc}
\end{small}
\end{center}
\end{subtable}

\begin{subtable}{\linewidth}
\begin{center}
\begin{small}
\begin{sc}
\begin{tabular}{lp{0.1\textwidth}<{\centering}p{0.1\textwidth}<{\centering}p{0.1\textwidth}<{\centering}p{0.1\textwidth}<{\centering}p{0.1\textwidth}<{\centering}c}
\toprule
\textbf{WALMART}       & BU             & TDHP           & TDFP          & OC    & TM             & Trainable (ours) \\
\midrule
mase - level 0  \hspace{3mm}   &   0.6812   &   0.8417   &   0.6808   &   0.6827   &   0.6795   &   0.6783  \\
mase - level 1   &   0.6269   &   0.6747   &   0.6274   &   0.6248   &   0.6223   &   0.6205  \\
mase - level 2   &   0.5795   &   0.5698   &   0.5698   &   0.5689   &   0.5722   &   0.5714  \\
\midrule
mlae - level 0   &   0.0726   &   0.0848   &   0.0724   &   0.0725   &   0.0724   &   0.0720  \\
mlae - level 1   &   0.1467   &   0.1542   &   0.1464   &   0.1459   &   0.1458   &   0.1453  \\
mlae - level 2   &   0.2980   &   0.2937   &   0.2937   &   0.2934   &   0.2949   &   0.2939  \\
\bottomrule
\end{tabular}
\end{sc}
\end{small}
\end{center}
\end{subtable}

\begin{subtable}{\linewidth}
\begin{center}
\begin{small}
\begin{sc}
\begin{tabular}{lp{0.1\textwidth}<{\centering}p{0.1\textwidth}<{\centering}p{0.1\textwidth}<{\centering}p{0.1\textwidth}<{\centering}p{0.1\textwidth}<{\centering}c}
\toprule
\textbf{AUSTRALIAN}      & BU             & TDHP           & TDFP          & OC    & TM             & Trainable (ours) \\
\midrule
mase - level 0 \hspace{3mm}  &   0.6542   &   0.7020   &   0.6905   &   0.7488   &   0.6388   &   0.6087  \\
mase - level 1   &   0.6479   &   0.7474   &   0.6859   &   0.7873   &   0.6206   &   0.5731  \\
mase - level 2   &   0.5769   &   0.7515   &   0.7549   &   0.7667   &   0.5978   &   0.4934  \\
mase - level 3   &   0.5160   &   0.6716   &   0.6716   &   0.6159   &   0.5377   &   0.4923  \\
\midrule
mlae - level 0   &   0.0601   &   0.0711   &   0.0649   &   0.0634   &   0.0587   &   0.0578  \\
mlae - level 1   &   0.0937   &   0.1224   &   0.1005   &   0.0993   &   0.0901   &   0.0875  \\
mlae - level 2   &   0.3629   &   0.4416   &   0.4614   &   0.4502   &   0.3828   &   0.3387  \\
mlae - level 3   &   0.8300   &   1.0047   &   1.0047   &   0.9300   &   0.8731   &   0.8120  \\
\bottomrule
\end{tabular}
\end{sc}
\end{small}
\end{center}
\end{subtable}

\begin{subtable}{\linewidth}
\begin{center}
\begin{small}
\begin{sc}
\begin{tabular}{lp{0.1\textwidth}<{\centering}p{0.1\textwidth}<{\centering}p{0.1\textwidth}<{\centering}p{0.1\textwidth}<{\centering}p{0.1\textwidth}<{\centering}c}
\toprule
\textbf{ELECTRICITY}     & BU             & TDHP           & TDFP          & OC    & TM             & Trainable (ours) \\
\midrule
mase - level 0 \hspace{3mm}  &   0.7193   &   0.9813   &   0.7403   &   3.9513   &   0.7203   &   0.7189  \\
mase - level 1   &   0.6944   &   0.9523   &   0.7083   &   2.3308   &   0.6923   &   0.6942  \\
mase - level 2   &   0.6510   &   0.8961   &   0.6620   &   0.7365   &   0.6561   &   0.6505  \\
mase - level 3   &   0.4439   &   0.4633   &   0.4633   &   0.4630   &   0.4555   &   0.4434  \\
\midrule
mlae - level 0   &   0.0376   &   0.0491   &   0.0382   &   0.0402   &   0.0380   &   0.0375  \\
mlae - level 1   &   0.0507   &   0.0665   &   0.0514   &   0.0531   &   0.0511   &   0.0505  \\
mlae - level 2   &   0.1887   &   0.2503   &   0.1915   &   0.1957   &   0.1915   &   0.1878  \\
mlae - level 3   &   2.0085   &   2.2147   &   2.2147   &   2.2129   &   2.0325   &   2.0025  \\
\bottomrule
\end{tabular}
\end{sc}
\end{small}
\end{center}
\end{subtable}
\end{table}

\bibliographystyle{unsrt}
\bibliography{bibliography}

\end{document}